\documentclass[conference]{IEEEtran}
\bstctlcite{IEEEexample:BSTcontrol}
\ifCLASSINFOpdf

\else

\fi

\usepackage{graphicx}
\usepackage{subcaption}
\usepackage{float}

\usepackage{amsmath,amssymb,epsfig}

\begin{document}
	
	\title{Spatio-Temporal RBF Neural Networks}
\author{\IEEEauthorblockN{Shujaat Khan\IEEEauthorrefmark{1}, Jawwad Ahmad\IEEEauthorrefmark{2}, Alishba Sadiq\IEEEauthorrefmark{3}, Imran Naseem\IEEEauthorrefmark{3} and Muhammad Moinuddin\IEEEauthorrefmark{4}}
		\IEEEauthorblockA{\IEEEauthorrefmark{1}Iqra University, Karachi.\\
            Email: shujaat@iqra.edu.pk\\
		  \IEEEauthorrefmark{2}Usman Institute of Technology,  Karachi-75300, Pakistan.\\
			Email: jawwad@uit.edu\\
			\IEEEauthorrefmark{3}COE, PAF-KIET, Karachi 75190, Pakistan.\\
			Email: \{alishba.sadiq, imrannaseem\}@pafkiet.edu.pk\\		    
		    \IEEEauthorrefmark{4}CEIES, King Abdulaziz University, Saudi Arabia.\\ 
			Email: mmsansari@kau.edu.sa}
			}

	\maketitle
	
\begin{abstract}
Herein, we propose a spatio-temporal extension of RBFNN for nonlinear system identification problem.  The proposed algorithm employs the concept of time-space orthogonality and separately models the dynamics and nonlinear complexities of the system.  The proposed RBF architecture is explored for the estimation of a highly nonlinear system and results are compared with the standard architecture for both the conventional and fractional gradient decent-based learning rules.  The spatio-temporal RBF is shown to perform better than the standard and fractional RBFNNs by achieving fast convergence and significantly reduced estimation error. 

\end{abstract}
	
	\IEEEpeerreviewmaketitle

\section{Introduction}
\label{intro}
Cybernetics and neural learning systems are becoming essential part of the society. They offer solution to several practical problems that occur in different areas ranging from medical sciences \cite{RAFP_Pred, ECMSRC, DNN_US}, to the control of complicated industrial processes.  With the advent of high power computing systems and deep network architectures enormous success has been achieved in diverse applications \cite{DNN_US}. However, deep learning typically demands very large amount of data  to extract the useful information from training samples. The need for high computational power and sufficiently large data-sets make implementation of deep architecture unrealistic for many simple applications. Whereas, the classical neural networks still provide reasonable performance with low computational complexity and small training data-set.

Artificial neural networks (ANNs) have the potential to model non-linear relationships easily, which is truly significant because most of the real-life input-output relationships are  non-linear in nature.  The simple architecture of the classical NNs if properly implemented can assure an appropriate performance therefore they are still in use for applications such as: data validation, risk management for the analysis and control of project risks, digit recognition for computerized bank check numbers reading, texture examination to detect microcalcifications on mammograms etc.  Modeling of engineering systems is an essential task, it is a primary step in system design and also helps in understanding the behaviour of the system for unknown scenarios.

Another important application is the fitness tracking of the systems with the help of redundant neural networks, the network helps in monitoring complex manufacturing machines.
A widely used simple neural network architecture is a radial basis function (RBF). The RBF neural networks give desirable results in most of the practical problems.  RBF neural networks are implemented in face recognition, it is also applied in pattern recognition due to the arbitrary nonlinear mapping characteristic.  Its application can also be found in the forecasting problems such as the time series forecasting because of the simple topological structure etc.  Recently, numerous modern RBF architectures have been introduced with the objective to provide improved performance by exploiting weight structures in RBFNN through Bayesian method.  In \cite{khan2016novel}, an adaptive kernel is proposed, the algorithm combines two widely used kernels to improve non-linear mapping of the signal in kernel space.

Spatio-temporal modeling of systems is becoming a significant area of research in various fields such as health-care, Epidemiology, medical imaging, ecosystem monitoring, business and operations etc.  Spatio-temporal analyses provides more advantages over uni-dimensional kernel space analysis.  Limited bandwidth and susceptibility to interference limits the use of wireless communication.  However, using temporal and spatial processing better trade-offs can be achieved, thus providing better performance.  Recently, a multi-object tracking algorithm is proposed with spatial-temporal information and trajectory of confidence.  Spatial-temporal correlation has been used to design a model which is more efficient and can deal with missed detection.

This motivated us to propose improvements in the architecture of RBF by utilizing the spatio-temporal (ST)-based signal processing. Thereby, an ST-RBF-NN for nonlinear system identification problem is presented.  The proposed algorithm utilizes the concept of time-space orthogonality and separately deals with the temporal signal(dynamics) and non-linearity(complexity) of the system.

The main contributions of our work are:
\begin{enumerate}
	\item A spatio-temporal extension of radial basis function (RBF) neural networks is proposed for nonlinear system identification problem.
	\item For the proposed architecture, an stochastic gradient descent-based weight update rule is derived.  
	\item  For performance evaluation, the proposed spatio-temporal RBFNN is implemented on nonlinear system identification and results are compared with the conventional and fractional order gradient descent-based \cite{khan2017novel,VPFLMS,RVSSFLMS,RVPFLMS,FCLMS,FLMF,CFLMScomments,mFLMScomments,wahab2019comments} variant of the RBFNN \cite{FRBF}.
\end{enumerate}
   Organization of this article is as follows. In section \ref{sec:proposed}, Spatio-temporal RBF neural network is discussed followed by an overview of RBF network architecture and derivation of the proposed method.  Algorithms are analyzed for the nonlinear system identification problem in section \ref{sec:results} and conclusion of results is discussed in section \ref{sec:conclusion}.

\section{The Spatio-Temporal RBF Neural Network} \label{sec:proposed}
\begin{figure}[h!]
	\begin{center}
		\centering
		\includegraphics[width=9cm]{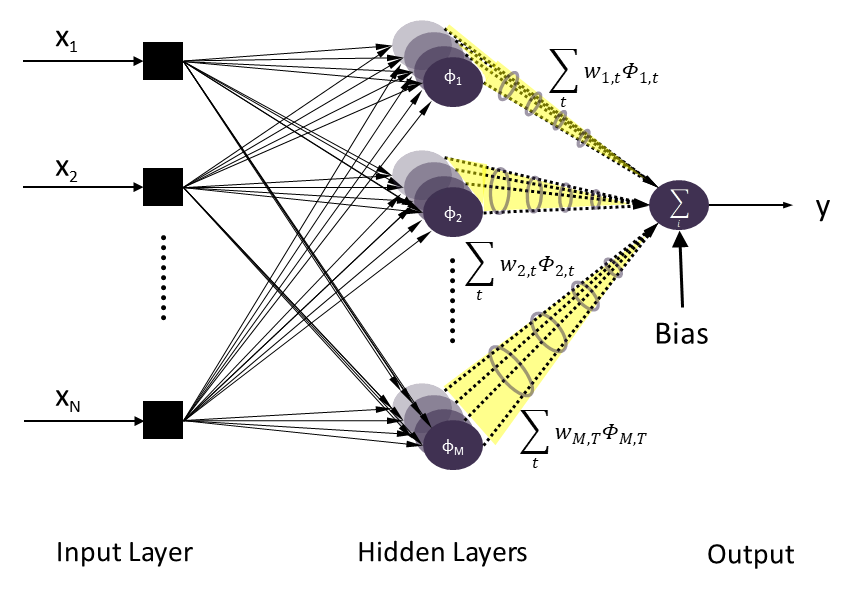}
	\end{center}
	\caption{Signal flow diagram of the spatio-temporal RBF neural network.}
	\label{rbfarch}
\end{figure}

Typically the RBFNN consists of 3 layers, including one input and one output layer. The non-linearity in RBFNN comes from a nonlinear hidden layer. However, in spatio-temporal processing of signal we also need temporal expansion of signal in kernel space as shown in Fig.\ref{rbfarch}.  For conventional architecture, consider an input vector $\mathbf{x} \in \mathbb{R}^{S}$ , then the final mapping of the RBFNN, $s:\mathbb{R}^{S}\rightarrow\mathbb{R}^{1}$, is given as

\begin{eqnarray}
y=\sum_{i=1}^{S}w_{i}\phi_{i}(\left\|\mathbf{x}-\mathbf{c}_{i}\right\|)+b,
\end{eqnarray} 
where $\mathbf{c}_i \in \mathbb{R}^{S}$ are the centers of the RBF network, $S$ is the number of neurons in the hidden layer, $w_i$ are the link weights that connect the nonlinear layer to output linear layer, the basis function is represented by the symbol $\phi_i$ and bias $b$ is added in the final output of neural network.  Generally for easiness only one output neuron is considered.

The output layer of RBF is linear, therefore to make non-linear decision boundary of the features linear We use the concept of Cover's theorem \cite{covers}, by which mapping from lower dimensional space to higher dimensional make the features space linearly separable in kernel space.  some popular mapping functions are as follows \cite{mapping}.\\
Multiquadrics:
\begin{equation}
  \phi_{i}(\|\mathbf{x-c}_{i}\|) = (\|\mathbf{x-c}_{i}\|^{2})^{\frac{1}{2}}  
\end{equation}
Inverse multiquadrics:
\begin{equation}
 \phi_{i}(\|\mathbf{x-c}_{i}\|) = \frac{1}{(\|\mathbf{x-c}_{i}\|^{2}+\zeta^{2})^{\frac{1}{2}}}
\end{equation}
Gaussian:
\begin{equation}\label{Gradial}
 \phi_{i}(\|\mathbf{x-c}_{i}\|)= \exp \left({\frac{-\|\mathbf{x-c}_{i}\|^{2}}{\sigma_{i}^{2}}} \right)
 \end{equation}
 where $\zeta$ is a nonzero positive constant and $\sigma$ is a standard deviation of Gaussian kernel.
\subsection{Gradient descent-based spatio-Temporal RBFNN}
For the training of proposed statio-temporal RBFNN, we employed the widely used least square method \cite{abbas2018topological} called gradient descent algorithm. Consider an RBFNN as shown in Fig.\ref{rbfarch}, the final mapping, at the $k$th learning iteration at a particular epoch, is given as:
\begin{eqnarray}
\label{map}
y(k)=\sum_{i=1}^{S}\sum_{t=1}^{T}w_{(i,t)}(k)\phi_{(i,t)}(\mathbf{x},\mathbf{c}_{(i,t)})+b(k),
\end{eqnarray} 
where $T$ is the truncated time, the bias $b(k)$ and synaptic weights $w_{(i,t)}(k)$ are updated at each iteration.  The cost function $\mathcal{E}(k)$ is defined as:
\begin{equation}
\label{cost}
\mathcal{E}(k)=\frac{1}{2}(d(k)-y(k))^{2} = \frac{1}{2}  e^2(k),
\end{equation}
The target signal at $k$th instant is given by  a symbol $d(k)$ whereas $e(k)$ is the immediate difference in estimated target  $e(k)=d(k)-y(k)$.

The conventional gradient descent based learning rule is given as:
\begin{equation}\label{eq:WeightConven}
w_{i}(k+1) = w_{i}(k) - \eta \nabla_{w_{i}} \mathcal{E}(k).
\end{equation}

We propose to use the spatio temporal extension of RBF, thereby equation \eqref{eq:WeightConven} can be given as:
\begin{equation}\label{eq:WeightConven1}
w_{(i,t)}(k+1) = w_{(i,t)}(k) - \eta \nabla_{w_{(i,t)}} \mathcal{E}(k).
\end{equation}
where $\eta$ is the step size, $w_{(i,t)}(k)$ and $w_{(i,t)}(k+1)$ are the current and updated weights.  
Evaluating the factor $-\nabla_{w_{(i,t)},}\mathcal{E}(k)$ by making use of the chain rule: 
\begin{equation}\label{Normal} -\nabla_{w_{(i,t)}} \mathcal{E}(k) =   -\frac{\partial \mathcal{E}(k)}{\partial e(k)} \times\frac{\partial e(k)}{\partial y(k)} \times \frac{\partial y(k)}{\partial w_{(i,t)}(k)} .
\end{equation}
After taking partial derivatives equation (\ref{Normal}) is simplified to:
\begin{equation}\label{Normal_Sim}
- \nabla_{w_{(i,t)}} \mathcal{E}(k) =   \phi_{(i,t)}(\mathbf{x},\mathbf{c}_{(i,t)}) e(k),
\end{equation}

using \eqref{Normal_Sim} equation \eqref{eq:WeightConven1} is turn out to be:		
\begin{equation}\label{weightf}
w_{(i,t)}(k+1) = w_{(i,t)}(k) + \eta  \phi_{(i,t)}(\mathbf{x},\mathbf{c}_{(i,t)}) e(k) .	
\end{equation}

In the same way the learning rule for $b(k)$ is given as:
\begin{equation}
b(k+1) = b(k) +  \eta e(k) .
\end{equation}

For complete weight vector $\mathbf{w}$ Eq \eqref{weightf} can be written as
\begin{equation}
\mathbf{w}(k+1) = \mathbf{w}(k) + \eta \; \mathbf{\phi}(\mathbf{x},\mathbf{c}) e(k).
\end{equation}

\section{Nonlinear System Identification using Proposed Spatio Temporal-RBFNN}\label{sec:results}
\begin{figure}[H]
	\begin{center}
		\centering
		\includegraphics[width=10cm]
		{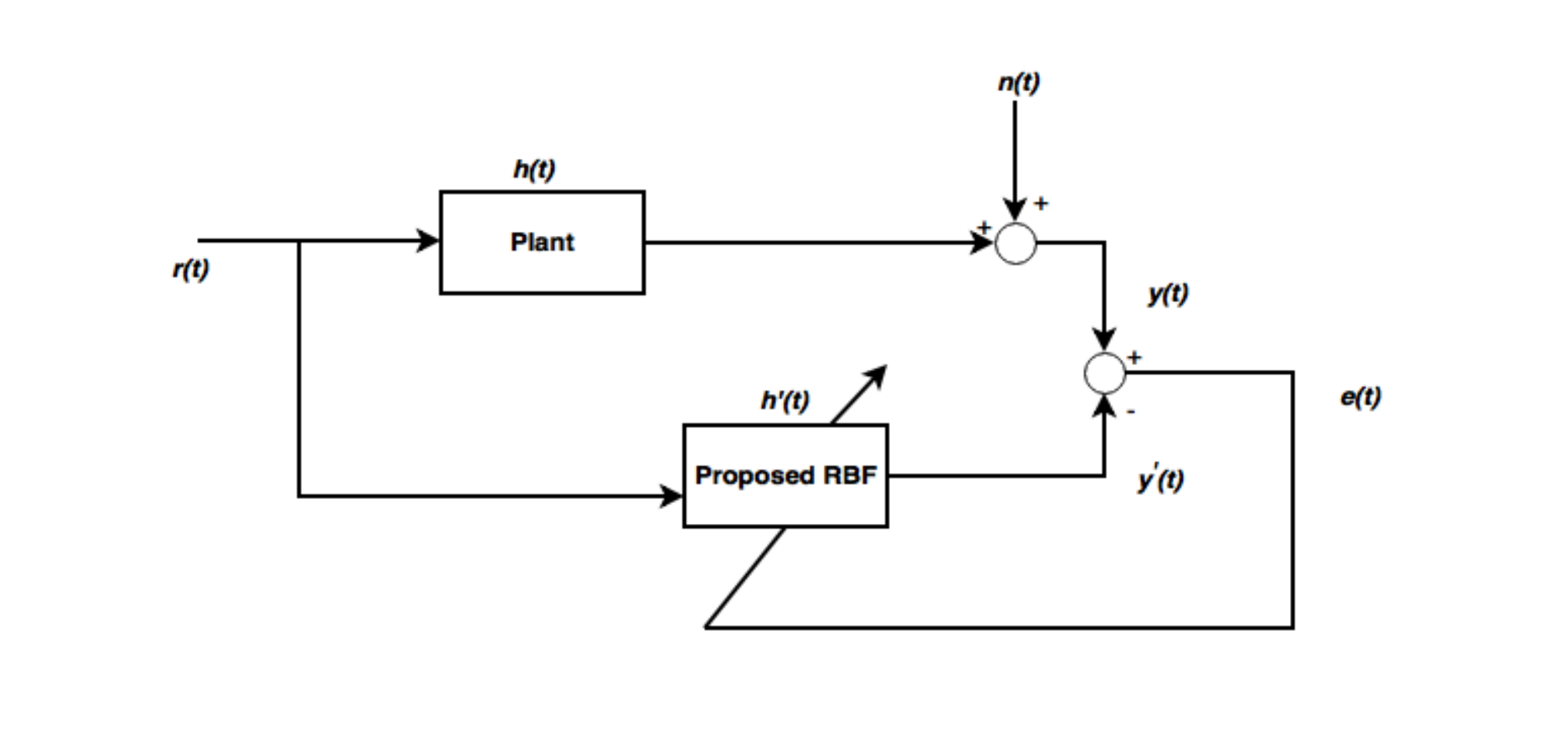}
	\end{center}
	\caption{Modelling of non-linear system using adaptive learning algorithm}
	\label{plant1}
\end{figure}

Nonlinear system identification is a rich topic and modeling of such system is difficult challenge. An iterative method can be used to estimate the computer model of a system by making use of instantaneous inputs and the corresponding outputs of the system.  Fig \ref{plant1}, shows the graphical model of the system identification problem.  Here, $h(t_{k})$ is defined as the coefficients of the system, while $\hat{h}(t_{k})$, $\hat{y}(t_{k})$ and $e(t_{k})$ are the estimated impulse response, estimated output and the estimated error respectively at time $t\rightarrow t_k$.   

For the evaluation of the proposed spatio-temporal RBF algorithm, a highly nonlinear system is considered:
\begin{multline}
y(t_{k})=q_1r(t_{k})+q_2r(t_{k-1})+q_3r(t_{k-2})+q_4[\cos(q_5 r(t_{k}))\\+exp^{-|r(t_{k})|}]+n(t_{k})
\end{multline}  
The input signal and target output are represented by the symbol $r(t_{k})$ and $y(t_{k})$ respectively, $q_i$s are the polynomial coefficients, $n(t_{k})$ noise source model considered to be $\mathcal{N}(0,\sigma_d^2)$.    For the training phase, the step function consists of $1000$ samples is taken as an input $r(t_{k})$ containing the first $250$ samples of value $1$  and the last $250$ samples were set to $-1$ and the sequence is repeated twice to make a full signal of length $1000$. In the testing phase the signal of length $200$ is used, hence increasing its frequency by $2.5 \times$ times. The experiment was executed on a noise level of $\sigma_{d}^{2} = 0.1$. System coefficients selected for simulations are: $q_1=2$, $q_2=-0.5$, $q_3=-0.1$, $q_4=-0.7$, and $q_5=3$.

\begin{figure}[H]
	\begin{center}
		\centering
		\includegraphics[width=8cm]
		{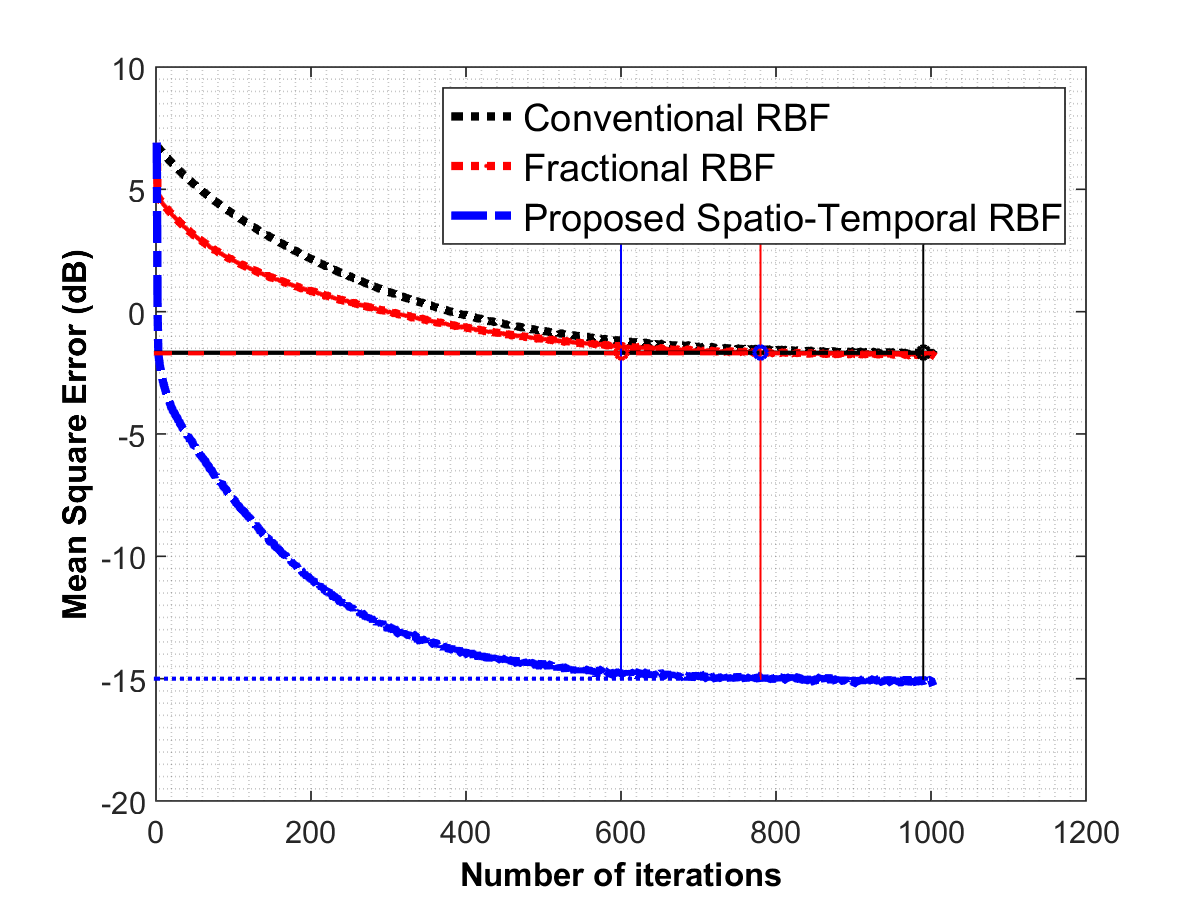}
	\end{center}
	\caption{MSE behavior of the conventional RBF, Fractional RBF and the proposed spatio-temporal RBF in the training phase.}
	\label{fig:MSE}
\end{figure}

For the simulations, the choice of neural network architecture and tuning parameters are as follows. We set the architecture such that the input layer consist of $3$ inputs i.e, $1$ current and $2$ previous values, $6$ neurons are selected and the centers are taken in the range from $-5$ to $5$ with a step size of $2$.  To obtain statistically significant results, we followed the Monte Carlos simulations protocol. Therefore all simulations were repeated by resetting the initial parameters with random values obtained from Gaussian distribution. $1000$ independent rounds were performed and averaged results of the experimental findings are reported.  

For performance evaluation, we compared our results with conventional and fractional RBFNN. For FRBFNN the value of convex combination weight $\alpha=0.5$ is chosen, the fractional order power $\nu$ is taken to be $0.9$.   
The values of step-sizes $\eta$ and $\eta_v$ were both selected as $2\times10^-5$.  While for the proposed spatio-temporal RBF $\eta$ is set to be half of the RBF and FRBF learning rates i.e., $1\times10^-2$.  The spread ($\sigma$) of the kernel which is Gaussian is selected to be $1$.  The training is performed by minimizing the mean squared error (MSE) cost function Eq \eqref{cost}. The learning behaviour is depicted in Fig. \ref{fig:MSE}.  The proposed Spatio-Temporal RBFNN produced lowest error i.e., $-15.1286$ dB while the RBFNN, and FRBFNN achieve the MSE of $-1.6813$ dB and $-1.7444$ dB, respectively.

\begin{figure}[H]
	\begin{center}
		\centering
		\includegraphics[width=8cm]
		{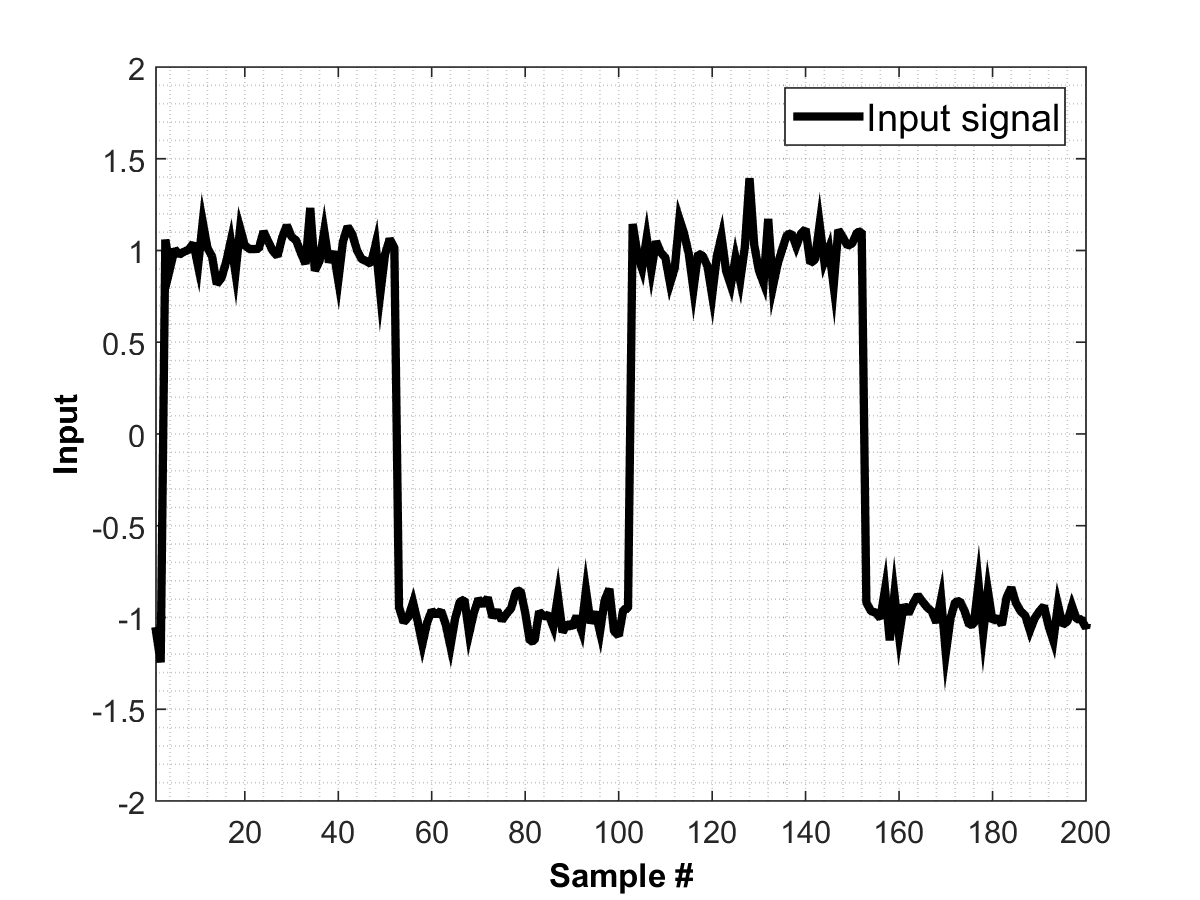}
	\end{center}
	\caption{Input signal.}
	\label{input}
\end{figure}

\begin{figure}[H]
	\begin{center}
		\centering
		\includegraphics[width=8cm]
		{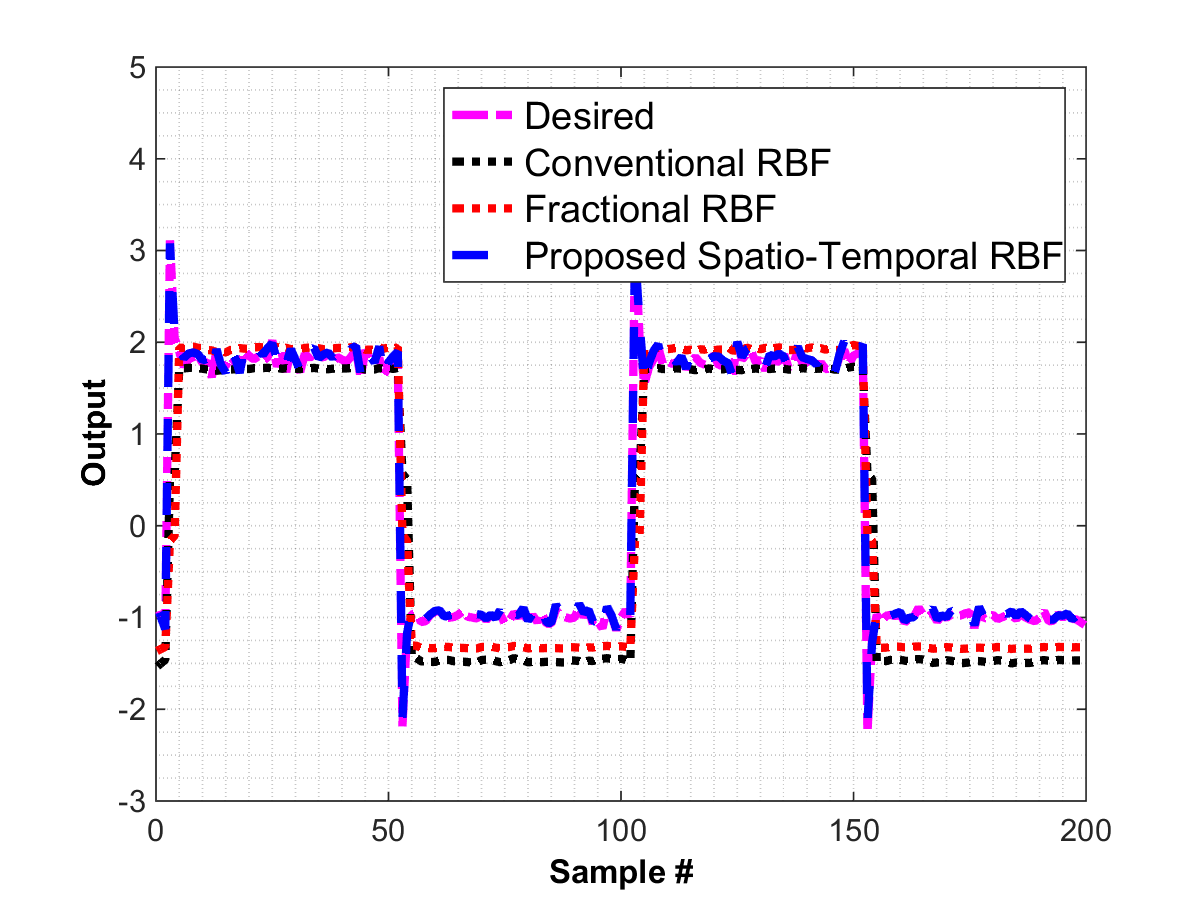}
	\end{center}
	\caption{Comparison of model and actual output.}
	\label{model_output}
\end{figure}

Fig. \ref{input}, shows the input signal of the neural network, Fig. \ref{model_output}, shows the comparison of the mean estimated output (test phase) of the proposed method with the desired response of the system and output of the RBFNN and FRBFNN.  From Fig. \ref{model_output} it can be observed that the target response and the output of the proposed spatio-temporal RBF shows the exact match as compared to the RBF and FRBF.  

\begin{figure}[H]
	\begin{center}
		\centering
		\includegraphics[width=8cm]
		{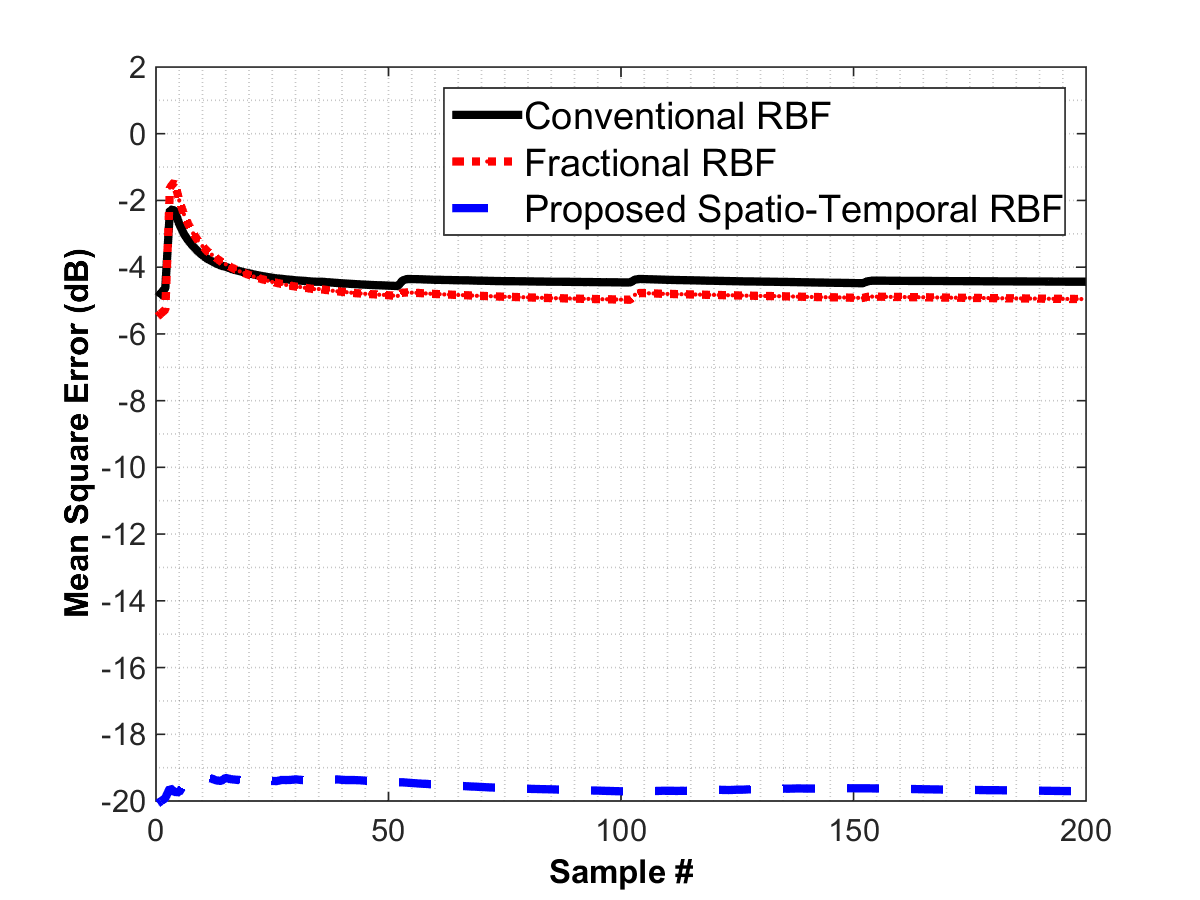}
	\end{center}
	\caption{MSE behaviour of the proposed spatio-temporal RBF with the conventional and fractional RBFNN in the testing phase.}
	\label{fig:test_mse}
\end{figure}

Finally the MSE in test phase is calclulated for $1000$ independent rounds and mean value of MSE is reported for different noisy test inputs. In Fig. \ref{fig:test_mse}, the output error of the proposed approach is compared to the output error of the conventional RBFNN and FRBFNN.  From Fig. \ref{fig:test_mse}, it can be observed that conventional RBF and the fractional RBF converges with the steady state error $-4.431$ dB and $-4.955$ dB, respectively. While the proposed spatio-temporal RBF neural network outperforms the conventional and fractional RBF by achieving the lowest steady state error of $-19.67$ dB. 

\section{Conclusion}\label{sec:conclusion}
For the nonlinear system identification, an RBF neural network with spatio-temporal extension is proposed.  The proposed algorithm is compared with standard and fractional variant of RBF.  Gradient descent-based learning algorithm is used for the training of adaptive weights parameters.  The MSE learning curves showed the overall best results for $1000$ independent runs.  Here we would like to point out that the proposed spatio-temporal RBFNN is an improvement in the architecture of standard RBF, its performance can be enhanced by incorporating better stochastic learning methods such as q-gradient descent \cite{EqLMS,qLMF}, further improvements can be made by adaptive learning of the optimal network connections, or through the introduction of dropouts probabilities etc. Code and supplementary material is available online \cite{STRBF_SYSTEM_CODE}.
	
	\bibliographystyle{IEEEtran}

\end{document}